\documentclass[11pt,letterpaper]{article}
\usepackage{emnlp2017}
\usepackage{times}
\usepackage{latexsym}

\usepackage{graphicx}
\usepackage{multirow}

\emnlpfinalcopy

\title{Exploiting Linguistic Resources for Neural Machine Translation\\Using Multi-task Learning}

 \author{Jan Niehues \and Eunah Cho \\
Institute for Anthropomatics and Robotics \\
KIT - Karlsruhe Institute of Technology, Germany \\
{\small \tt firstname.lastname@kit.edu} }

\date{}

\begin{document}

\maketitle

\begin{abstract}

Linguistic resources such as part-of-speech (POS) tags have been extensively used in statistical machine translation (SMT) frameworks and have yielded better performances. 
However, usage of such linguistic annotations in neural machine translation (NMT) systems has been left under-explored. 

In this work, we show that multi-task learning is a successful and a easy approach to introduce an additional knowledge into an end-to-end neural attentional model. 
By jointly training several natural language processing (NLP) tasks in one system, we are able to leverage common information and improve the performance of the individual task. 

We analyze the impact of three design decisions in multi-task learning: the tasks used in training, the training schedule, and the degree of parameter sharing across the tasks, which is defined by the network architecture. 
The experiments are conducted for an German to English translation task. 
As additional linguistic resources, we exploit POS information and named-entities (NE). 
Experiments show that the translation quality can be improved by up to 1.5 BLEU points under the low-resource condition. 
The performance of the POS tagger is also improved using the multi-task learning scheme. 

\end{abstract}

\section{Introduction}
\label{intro} 

Recently, there has been a dramatic change in the state-of-the-art techniques for machine translation (MT). 
In a traditional method, often the best performance is achieved by using a complicated combination of several statistical models, which are individually trained. 
For example, POS information was shown to be very helpful to model word reordering between languages, as shown in \citet{Niehues2009}. 
While the recent development of end-to-end trained neural models \cite{Bahdanau2014} showed significant gains over traditional approaches, they are often trained only on the parallel data in an end-to-end fashion. 
In most cases, therefore, they do not facilitate other knowledge sources. 

When parallel data is sparse, exploiting other knowledge sources can be crucial for performance. Two techniques to integrate the additional resources are well studied. 
In one technique, we train a tool on the additional resources (e.g. POS tagger) and then annotate the parallel data using this tool. 
This technique has been applied extensively in SMT systems (e.g. \newcite{Niehues2009}) as well as in some NMT systems (e.g. \newcite{Sennrich2016Factor}). 
The second technique would be to  use the annotated data directly to train the model. 

The goal of this work is to integrate the additional linguistic resources directly into neural models, in order to achieve better performance. 
To do so, we build a multi-task model and train several NLP tasks jointly. 

We use an attention-based sequence-to-sequence model for all tasks. 
Experiments show that we are able to improve the performance on the German to English machine translation task measured in BLEU, BEER and CharacTER. 
Furthermore, we analyze three important decisions when designing multi-task models. 
First, we investigated the influence of secondary tasks. 
Also, we analyze the influence of training schedule, e.g. whether we need to adjust it in order to get the best performance on the target task. 
And finally, we evaluated the amount of parameter sharing enforced by different model architectures.

The main contributions of this paper are (1) that we show multi-task learning is possible within attention-based sequence-to-sequence models, which are state-of-the-art in machine translation and (2) that we analyze the influence of three main design decisions.

\section{Related Work}

Motivated by the success of using features learned from linguistic resources in various NLP tasks, there have been several approaches including external information into neural network-based systems. 

The POS-based information has been integrated for language models in \citet{Wu2012,Niehues2016}. In the neural machine translation, using additional word factors like POS-tags has shown to be beneficial \cite{Sennrich2016Factor}. 

The initial approach for multi-task learning for neural networks was presented in \citet{Collobert2011}. 
The authors used convolutional and feed forward networks for several tasks such as semantic parsing and POS tagging. This idea was extended to sequence to sequence models in \citet{Luong2015}.

A special case of multi-task learning for attention based models has been explored. In multi-lingual machine translation, for example, the tasks are still machine translation tasks but they need to consider different language pairs. In this case, a system with an individual encoder and decoder \cite{Firat2016} as well as a system with a shared encoder-decoder \cite{Ha2016, Johnson2016} has been proposed.

\subsection{Attention Models}

Recently, state-of-the art performance in machine translation was significantly improved by using neural machine translation. 
In this approach, a recurrent neural network (RNN)-based encoder-decoder architecture is used to transform the source sentence into the target sentence.

In the encoder, an RNN is used to encode the source sentence into a fixed size of continuous space representation by inserting the source sentence word-by-word into the network. 
First, source words are encoded into a one-hot encoding. Then a linear transformation of this into a continuous space, referred to as word embeddings, is learned. 
An RNN model will learn the source sentence representation over these word embeddings. 
In a second step, the decoder is initialized by the representation of the source sentence and is then generating the target sequence one word after the other using the last generated word as input for the RNN. In order to get the output probability at each target position, a softmax layer that get the hidden state of the RNN as input is used \cite{Sutskever2014}. 

The main drawback of this approach is that the whole source sentence has to be stored in a fixed-size context vector. 
To overcome this problem, \citet{Bahdanau2014} introduced the soft attention mechanism. Instead of only considering the last state of the encoder RNN, they use a weighted sum of all hidden states. Using these weights, the model is able to put attention on different parts of the source sentence depending on the current status of the decoder RNN. In addition, they extended the encoder RNN to a bi-directional one to be able to get information from the whole sentence at every position of the encoder RNN.
A detailed description of the NMT framework can be found in \citet{Bahdanau2014}.

\section{Multi-task Learning}

In a traditional NLP pipeline, a named entity recognition or machine translation system employ POS information by using the POS tags as additional features. For example, the system will learn that the probability of a word being a named entity is higher if the word is marked as a noun. First, a POS tagger is used to annotate the input data. 
Combining the statistical models used for POS tagging and named entity recognition might not be straightforward. 

Recent advances in deep learning approaches, e.g. CNN or RNN-based models \cite{Labeau2015}, made it straightforward to use very similar techniques throughout different NLP tasks. 
Therefore, there are new methods to combine the tasks. 
Instead of using the output of a model as input for another one, for example, we can build one model for all tasks. The model is then automatically able to learn to share as much information across the tasks as necessary. 

For building a model that can learn three NLP tasks, we use the attention-based encoder-decoder model, which is a standard in state-of-the-art machine translation systems. The two non-MT tasks can also be modeled by converting them into a translation problem. 
Instead of translating the source words into the target language, we \textit{translate} the words into labels, either POS-tags or NE-labels.

In this work, we study several crucial design aspects when applying attention-based encoder-decoder model for a multi-task learning scenario. 
First, we consider different architectures of the network in order to assess how much parameter sharing is useful between the tasks. 
In general, sharing more information across the tasks is preferred. 
However, if the tasks differ from each other greatly, it might be helpful to restrict the degree of sharing. In addition, the training schedule of each task has to be addressed. 
While all three tasks are handled as a form of translation, certain distinctions and special processes needed to be asserted. 
In Section \ref{tlength} we address this issue.

\subsection{Architecture}

\begin{figure*}
\centering
\caption{\label{figOverview}Overview on the different architectures used for multi-task learning}
\includegraphics[width=0.65\textwidth]{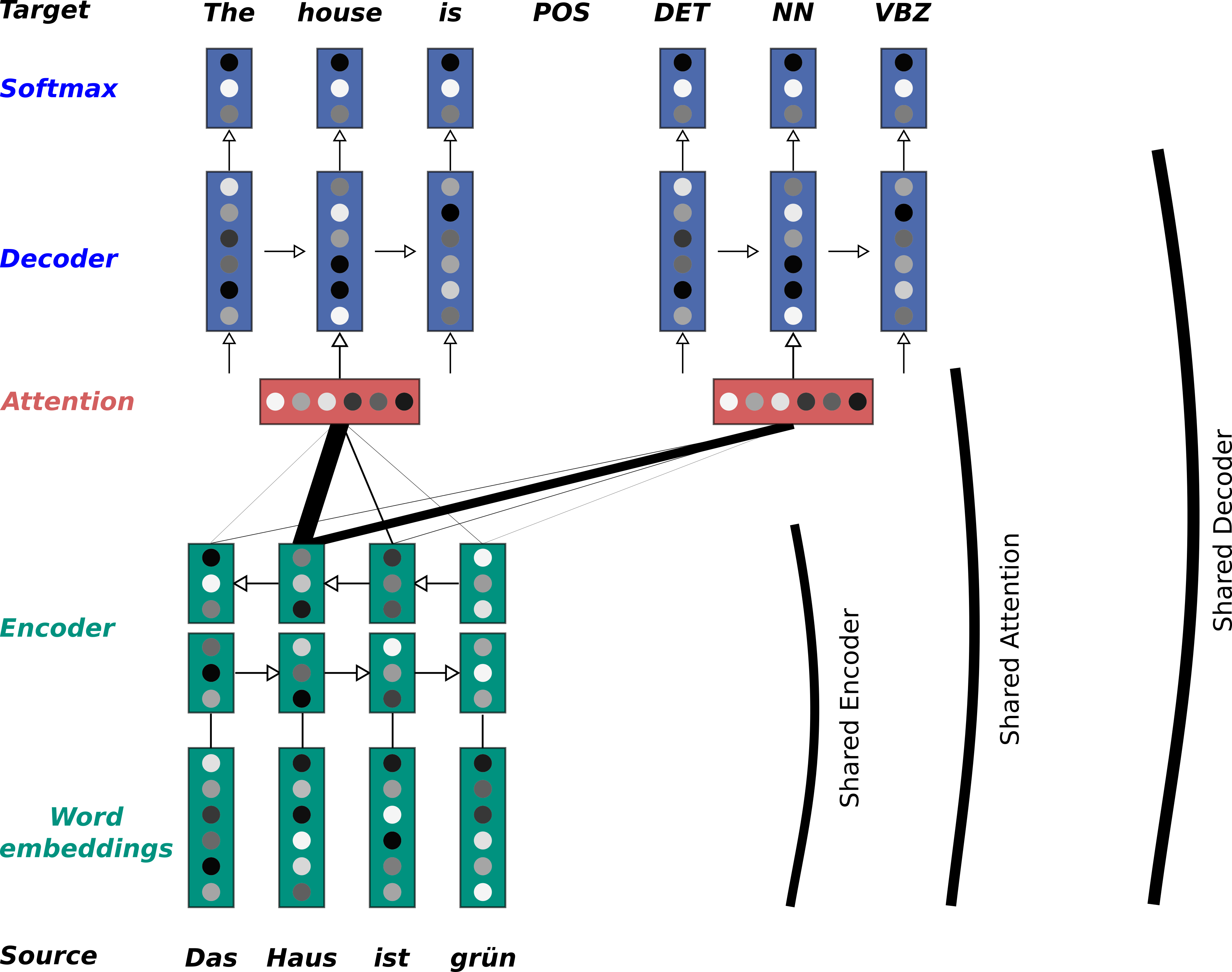}
\end{figure*}

The general attentional encoder-decoder model consists of three main parts: the encoder $E$, the attention model $A$ and the decoder $D$. Figure \ref{figOverview} gives an overview of this layout. 

Our baseline considers the scenario where we have separate models for each task. 
Therefore, all three parts (encoder, attention model, and decoder) stand separately for each task. 
We will have nine components $E_{MT}$, $E_{POS}$, $E_{NE}$, $A_{MT}$, $A_{POS}$, $A_{NE}$, $D_{MT}$, $D_{POS}$, $D_{NE}$ in total. 

The one main design decision for a multi-task learning architecture is the degree of sharing across the tasks. 
Motivated by architectures proposed for multi-lingual machine translation \cite{Dong2015,Firat2016NAACL, Ha2016}, we analyze the impact of different degrees of sharing in the output quality.  
When sharing more parameters between the tasks, the models are able to learn more from the training data of other tasks. 
If the tasks are very distant, on the other hand, it might be harmful to share the parameters.

\paragraph{Shared encoder \textit{(shrd Enc)}}

One promising way is to share components that handle the same type of data. Since all our tasks share English as input here is the encoder. 

In this architecture, we therefore use one encoder for all tasks. 
This is the minimal degree of sharing we consider in our experiments. 
A common encoder $E_{ALL}$ is used for all tasks, but 
separate attention models $A_{MT}$, $A_{POS}$, $A_{NE}$ and decoders $D_{MT}$, $D_{POS}$, $D_{NE}$ are used.

\paragraph{Shared attention \textit{(shrd Att)}}

The next component is the attention model which connects the encoder and decoder. 
While the output should be different for the addressed tasks, the type of input is the same. 
Therefore, it might be helpful to share more information between the models. 

In a second architecture, we also share the attention model in addition to the encoder. 
So in this setup, we have one encoder $E_{ALL}$, one attention model $A_{ALL}$ and three decoder $D_{MT}$, $D_{POS}$, $D_{NE}$.

\paragraph{Shared decoder \textit{(shrd Dec)}} 

Finally, we explore whether it is possible to share all information across the tasks and let the model learn how to represent the different tasks. 
Thus, in this scheme, we aim to share the decoder partially. 
The only thing that is not shared is the final softmax layer. 

In this architecture, the decoder RNN has to model the generation of target words as well as that of labels. 
Therefore, we have only one encoder $E_{ALL}$, one attention model $A_{ALL}$ and one decoder $D_{ALL}$. 
In the decoder, however, we have separated output layers for each task. 

Figure \ref{figOverview} depicts which layers are shared depending on the architecture.

\subsection{Training Schedule}
\label{training}

In this section, we discuss the influence of the training schedule on the quality of the model. 

Throughout our experiments we used a mini-batch size of 512 tokens.
The weight updates were determined using the Adam algorithm.

The training has to be adapted to the multi-task scenario. 
The main decision is how to present the training examples to the training algorithm. 
We only consider one task in each mini-batch. Although the model structure is the same for all tasks, the models for the individual tasks have different weights. Therefore, parallelization on the GPU would be less efficient when using different tasks within one batch. 
In order to train our model on all tasks in parallel, we randomly shuffle the mini-batches from all tasks. This is our default training schedule.
One issue in the multi-task scenario is that the data size might vary. 
In this case, the model will mainly concentrate on the task with the most data and not achieve the best performance on each task.

This challenge is strongly related with the problem of domain adaptation in machine translation, where a large out-of-domain data is available but only a small amount of in-domain data. 
For this scenario, first training on all data and then fine-tuning on the in-domain data was very successful \cite{Lavergne2011,Cho2016IWSLT}. 
Therefore, we adapt this approach to the multi-task scenario. 
In this case, we first trained the model on all tasks and then continued training only on the main task. We will refer to this training schedule as adapted.

\subsection{Target Length}
\label{tlength}

While all tasks are modeled as a translation problem in this work, the nature of each task is largely different. 
One main difference between the translation task and the other two tasks is the length of the target sequence. 
While it is unknown in the translation task, it is known and fixed for the other two cases. 
During training this does not matter as the target sequence is given. 
For testing the system, however, this issue is crucial to address.

In our initial experiment, it was shown that the POS tagger was able to learn the correct target length in most of the cases. For some sentences, however, the estimated target length was not correct. 
Therefore, the prior knowledge of sequence length is used during decoding so that label sequences are generated with the correct target length. 
It is worth to mention that the desired length of the labels is not exactly the length of the input to the model itself. Our model uses inputs with subwords units generated by byte-pair encoding \cite{Sennrich2016}.

\section{Experimental Setup}

We conduct experiments using the multi-task approach on three different tasks: machine translation from German to English, German fine-grained POS tagging and German NE tagging. 
As briefly mentioned in Section \ref{intro}, multi-task approach can be helpful when data is sparse. 
In order to simulate this, we deploy only German to English TED data for the translation task.

\subsection{Data}

For the translation task, we used 4M tokens of the WIT corpus \cite{Cettolo2012WIT} for German to English as training data. 
We used \textit{dev2010} for validation and \textit{tst2013} and \textit{tst2014} for testing, provided by the IWSLT. 
We only used training examples shorter than 60 words per sentence. 

The POS tagger was trained on 720K tokens the Tiger Corpus \cite{Brants2004}. This corpus contains German newspaper text. Consequently, it is out-of-domain data for the machine translation task.
The development and the test data are also from this corpus. 
The POS tag set consists of 54 tags and the fine-grained POS tags with morphological annotations has 774 labels.

Finally, we trained the German named-entity tagger on 450K tokens of the GermEval 2014 NER Shared Task data \cite{Benikova2014}. 
The corpus is extracted from Wikipedia and the training data consists of 24K sentences.

We preprocess the parallel data by tokenizing and true-casing. 
In addition, we trained a byte-pair encoding \cite{Sennrich2016} with 40K subwords on the source and target side of the TED corpus jointly. 
We then applied the subwords to all German and English corpora. 

\subsection{System Architecture}

For all our experiments, we use an attentional encoder-decoder model. 
The baseline systems use this architecture as well. 
The encoder uses word embeddings of size 256 and a bidirectional LSTM \cite{Hochreiter1997,Schuster1997} with 256 hidden layers for each direction. 
For the attention, we use a multi-layer perceptron with 512 hidden units and tanh activation function. 
The decoder uses conditional GRU units with 512 hidden units. 
The models are all trained with Adam, where we restarted the algorithm twice and early stopping is applied using log-likelihood of the concatenated validation sets from the considered tasks. For the adapted schedule, Adam is started once again when training only on the target task.
The model is implemented in lamtram \cite{Neubig2015lamtram}\footnote{The extension to handle multi-task training can be downloaded \textit{https://github.com/isl-mt/lamtram}}.

\subsection{Evaluation}

The machine translation output is evaluated with BLEU \cite{Papineni2002}, BEER \cite{Stanojevic2014} and CharacTER \cite{Wang2016}. 
For the POS tags, we report error rates on the small label set as well as on the large label set. 

\section{Results}

In this section, we present the results from our experiments and analysis. 

\subsection{Initial experiments on the architecture} 
\label{initial} 
The results of the initial experiments on the machine translation tasks are shown in Table \ref{MTresults}. 
The table displays the performance on the validation set and on both test sets. For all experiments, we first show the BLEU score, then the BEER score and finally the characTER. 

 \begin{table*}[htb]
  \begin{center}
   \begin{tabular}{l|c|c|c|c} \hline 
   
    Task(s) & Arch. & Valid & \multicolumn{2}{c}{Test} \\  \cline{4-5}
     & & dev 2010 &tst2013&tst2014\\ \hline 
MT & - & 29.91/62.16/51.06 & 30.85/62.27/51.16 & 26.12/58.73/55.17 \\  \hline 
\multirow{3}{*}{POS + MT} & shrd Enc  & 30.62/62.77/48.35 & 31.97/62.72/49.69 & 27.08/58.99/54.50 \\ 
 & shrd Att & 30.51/62.27/49.09 & 31.76/62.68/49.59 & 26.86/58.84/53.88 \\ 
  & shrd Dec & 30.36/62.34/49.28 & 31.26/62.31/50.35 & 26.52/58.48/54.00 \\ \hline
 Adapted NE + POS + MT & shrd Enc & 30.70/62.96/48.60 & 32.30/63.25/49.22 & 27.78/59.74/53.49\\

 \hline
   \end{tabular}
 \caption{\label{MTresults} Results of multi-task learning architectures on the machine translation task (BLEU/BEER/characTER)}
  \end{center}
 \end{table*}

First, we show the results of the baseline neural MT system trained on the parallel data (single task). 
As mentioned in the beginning, we simulated a low-resource condition in these experiments by only using the data from TED, which are roughly 185K sentences.

We evaluated models that are trained both on the translation and POS tagging task. 
Although the POS data is out-of-domain and significantly smaller than the parallel training data for the translation task (ca. 20\% of the size), we see improvements for all three architectures consistently in three metrics. The BLEU scores is improved by more than 1 point and the characTER is reduced by more than 1.5 points. The BEER metric score is improved by more than a half point on both sets.

In a more detailed look at this task, we see that the model sharing the most (\textit{shrd Dec}) performs better than the baseline, but worse than the other two. 
Therefore, we can conclude that it is helpful to separate the tasks when the components work on different types of data. 
Whether it is helpful to share the attention layer (\textit{shrd Att}) or not (\textit{shrd Enc}) is not clear from this experiment. 
Therefore, we concentrate on these two architectures in the following experiments.

\subsection{Impact of design decisions} 

Following the initial experiment, we address the following three design questions:

\begin{itemize}
\item What kind of influence does the secondary task have?
\item How do the different architectures perform?
\item Do we need to adapt the training schedule?
\end{itemize}

In order to clarify the impact of the three hyper-parameters (the architectures, the tasks and the training) we performed experiments based on possible combinations. 
We used two most promising architectures, \textit{shrd Enc} and \textit{shrd Att} as discussed in Section \ref{initial}. 
We use three task combinations, \textit{POS+MT}, \textit{NE+MT} and \textit{NE+POS+MT}.
Two training strategies are applied with and without adaptation as described in Section \ref{training}. 
These 12 systems are evaluated on the two test sets using three different metrics. 
Consequently, in total we have 72 measurements for the 12 systems.

Since a first view on the results did not clearly reveal a best performing system, 
we conducted a more detailed analysis by averaging the results over several configurations. 
First, we analyze the influence of adapting the training schedule by fine-tuning on the MT task. 
Out of the 12 systems, six systems used an adapted training schedule. 

 \begin{table*}[htb]
  \begin{center}
   \begin{tabular}{l|c|c|c} \hline 
Systems & Default Schedule & Adapted Schedule & Adapted better \\ \hline 
All & 29.48/60.89/52.05 & 29.89/61.08/51.64 & 25/36 \\ \hline
shrd Enc & 29.34/60.85/52.31 & 30.00/61.25/51.50 & 17/18 \\
shrd Att & 29.62/61.93/51.78 & 29.78/60.93/51.79 & 8/18 \\  \hline
POS + MT & 29.41/60.81/51.92 & 29.78/61.00/51.90 & 8/12\\
NE + MT & 29.60/61.00/51.76 & 29.79/60.96/51.77 & 5/12\\
NE + POS + MT & 29.42/60.87/52.46 & 30.09/61.46/51.25 & 12/12\\
 \hline 
   \end{tabular}
 \caption{\label{MTmeta} Impact of the training schedule in the machine translation task (BLEU/BEER/characTER)}
  \end{center}
 \end{table*}

As shown in the first line of Table \ref{MTmeta} (\textit{All}), when averaging over the six systems using the adapted training schedule and tested both test sets, 
we see improvements in all considered metrics compare to the systems using the default training schedule. 
The BLEU score improved by 0.4 BLEU points, BEER by 0.2 and characTER by 0.4. 
Furthermore, we compared each of the 36 measurements using the adapted schedule with the corresponding measurement using the default training schedule. 
Thus, the scores are calculated on same test set, based on the same metric. The model differs whether it is trained using the default or adapted training schedule. 
How often the system with an adapted schedule performs better is shown in the last column of Table \ref{MTmeta}. 
When directly comparing these systems, in 25 out of 36 cases the ones with the adapted schedule perform better. 

We analyzed the influence of the architecture as well as tasks considered in training in the same way. 
The influence of both aspects, however, was not as clear as the one from the training schedule. 
In order to get a deeper understanding, we analyzed in which cases it is more helpful to adapt the training schedule.
As a first step, we looked at the correlation between the training schedule and the two different architectures. 
The results are shown in the next lines of Table \ref{MTmeta}.

Compared to the systems using the \textit{shrd Enc} layout, we observe even bigger improvements when applying the adapted schedule. 
The averaged BLEU score is improved by 0.7 BLEU points. 
Furthermore, the system with the adapted training schedule performs better, in almost all cases. 
For the \textit{shrd Att} model, in contrast, we gain nearly no improvements from the adapted schedule. 
We also observed that the system with the default schedule performs better in 10 out of 18 cases. 

One reason for this can be that the default training schedule may not perform as well any more when only a few parameters are observed in every batch.  
In this case, continuing and concentrating on one task seems to be very important.

In addition, we evaluate the correlation between the tasks involved and the training schedule. 
The results are shown in the same table. 
The adapted training schedule has no effect when training on named entities and machine translation. 
The effect when training on POS tagging and MT is also relatively small.  
When training the three tasks together, however, the system with an adapted schedule performs always better than the system with the default one. The average BLEU is improved by 0.7. The BEER score and characTER are also improved by 0.5 and 1.2 points. 

Inspired by the results, we build the adapted \textit{shrd Enc} model trained on all three tasks, as shown in Table \ref{MTresults}. 
This model improved the performance by 1.5 BLEU points over the baseline system. 
Also the BEER score is improved by 1 and the characTER score reduced by 1.8 to 2 points. 

\subsection{POS Tagging Performance}

In addition to the results on the task of translation, we also evaluated the performance on the task of POS tagging. The results are shown in Table \ref{POSresults}. 

 \begin{table*}[!htb]
  \begin{center}
   \begin{tabular}{l|c|c|c|c|c} \hline 
   
    Task(s) & Model & \multicolumn{2}{c|}{Default schedule} &  \multicolumn{2}{c}{Adaptation schedule }  \\ \cline{3-6}
    & & Valid & Test & Valid & Test \\ \hline 
POS & - &  5.49/11.36 & 10.13/17.27 & - & - \\ \hline  
\multirow{3}{*}{POS + MT}& shrd Enc & 3.99/9.98 & 7.55/14.98 & 3.57/8.82 & 6.24/13.24 \\
& shrd Att & 3.86/9.55 &  6.98/14.17& 3.16/8.23 & 5.52/12.25 \\ 
 & shrd Dec & 3.57/9.28 & 7.40/14.62 & 3.53/8.94 & 5.81/12.56 \\ \hline 
 \multirow{2}{*}{NE + POS + MT} & shrd Enc & 3.42/9.00 & 5.86/12.87 & 3.00/8.00 &  5.06/11.62\\
& shrd Att & 3.08/8.45 & 6.23/13.28 & 2.78/7.87 & 5.49/12.10\\

 \hline 
   \end{tabular}
 \caption{\label{POSresults} Results of different multi-task architectures on the POS task}
  \end{center}
 \end{table*}

For the validation and test data, we show the error rate on the small tag sets as well as the error rate on the morpho-syntactic tag set. 
In the table, we always first show the results for the small test set. 

The baseline system trained only on the Tiger corpus achieves an error rate of 5.49, for the POS tags in the validation set. 
For the morpho-syntactic tag of the validation set, it achieves 11.36. 
The performance on the test data is 10.13 and 17.27 for both tag sets. 
In all systems we used one system the generate the both tag sets. The small tags were evaluated by removing the morhpo-syntactic information from the output

It is clear that all models outperform the baseline. 
It seems to be very helpful for the POS task to jointly train the model along with the translation task. 
The MT data is significantly larger than the POS data, which is beneficial for this task. 

A more detailed look shows that model adaptation is beneficial for a good performance. 
In all cases the performance is improved by adapting the model to the POS task. 
Therefore, when the data of the main task is small compared to the overall training data, adapting on the main task is even more important.

Furthermore, we see improvements when using a third task in all cases. Facilitating this combination of tasks is also helpful for POS tagging. 

As we observed in the MT task, the impact and differences brought from each architecture are not huge. 
The architectures considered in this work perform similar. 
Even the system sharing all components achieves a comparable performance on this task. 

The best performing model, however, is the \textit{shred Enc} model, trained on all three tasks and adapted to the task. 
This model achieved an error of 5.06 on the small tag set. Compared to the baseline performance of 10.13, we can see that the error rate is halved. 
On the fine-grained tag set, we see an improvement from 17.27 to 11.62, which is a more than 30\% reduction in error rate.

\subsection{Analysis and Examples}

In order to show the influence of the other tasks, we show translation examples in Table \ref{Example}.
For the examples we use the multi-task system trained on all three tasks with the \textit{shrd Enc} architecture.

 \begin{table*}[!htb]
  \begin{center}
   \begin{tabular}{l|p{13.5cm}} \hline 
German & sie ist kein Geburtsfehler. \\
Reference & it's not a birth defect.  \\
Baseline & she's not born. \\
Multi-task & it's not a birth error. \\ \hline
German & das bedeutet, dass 8 von 10 Entscheidungen... \\
Reference & that means that eight out of 10 of the decisions... \\
Baseline & that means that eight of 10 of 10 choices... \\
Multi-task & that means that eight of 10 decisions... \\ \hline
German & ...[``Benjamin Franklin'' von Walter Isaacson][``John Adams'' von David McCullough]... \\ 
Reference & ...[``Benjamin Franklin'' by Walter Isaacson][``John Adams'' by David McCullough]... \\
Baseline & ...[Benjamin Franklin, from Walter Franklin''][The ``John Adams'']... \\
Multi-task & ...[``Benjamin Franklin'' from Walter Isaacson],[``John Adams''  from David McCullough... \\ \hline
German & darum habe ich infantile Zerebralparese,  ... \\
Reference & as a result, I have cerebral palsy,  \\
Baseline & that's why I have the infantile, \\
Multi-task & I have infantile cerebral palsy, \\  \hline
German & Prousts Freunde h\"{a}tten das Land verlassen m\"{u}ssen, ..  \\
Reference & you know, Proust's boyfriends would have to leave the country ... \\
Baseline & Prolled friends had to have left the country  ... \\
Multi-task & Prouless friends have to leave the country  ... \\

 \hline 
   \end{tabular}
 \caption{\label{Example} Translation examples}
  \end{center}
 \end{table*}

A common problem of many neural MT systems is that they do not translate parts of the source sentence, or that parts of the source sentence are translated twice. 
The baseline system suffers from this, as shown in the first two examples. 
The translation of the multi-task system is improved compared to the baseline in several aspects. 
In the first example, the baseline system is not translating the German compound \textit{Geburtsfehler} into \textit{birth defect} correctly, but into \textit{birth}. 
Although the multi-task system does not generate the translation that exactly matches the reference the translation is understandable. 
In the second example, the phrase \textit{of 10} is not repeated. One explanation for this could be that the additional information from the POS data leads to a better encoding of the structure of the source sentence.

The influence of the named-entity training examples on the translation quality is clearer. 
In several cases, the model is able to handle named entities better. 
As shown in the third and fourth example, the NMT system is not able to copy a named entity from the source to the target, nor to translate rare words. 
In the third example, the baseline system is not able to generate the correct last name of the first author \textit{Isaacson}, 
but is generating the last name from the book title. 
In the second part of the example, the baseline system completely deletes the author. 
In contrast, the multi-task system is able to generate the correct sequence. 
In the fourth example the multi-task example is able to translate \textit{Zerebralparese (cerebral palsy)}, while the baseline system is not able to do it.

We would like to note that as shown in the last example, there are also several cases where the NMT system is not able to translate names or rare words correctly.

\section{Conclusion}

In this paper we proposed the use of multi-task learning for attention-based encoder-decoder models in order to exploit linguistic resourced for NMT. 
By training the models not only on the machine translation task, but also on other NLP tasks, we yielded clear improvements on the translation performance. 
Results show that multi-task learning improves the translation up to 1.5 BLEU points and 2 characTER points. 
As a by product, we were also able to improved the performance of the POS tagging by 30\% to 50\% relatively. 
This is especially helpful since data annotation for many NLP tasks is very time-consuming and expensive.  
It suggests that multi-task learning is a promising approach to exploit any linguistic annotated data, which is especially important if we have a low-resource condition.

We addressed the influence of three design decisions: the involved tasks, the training schedule and the architecture of the model. The largest influence on the final performance was given by the training schedule . By adapting the system on the individual tasks, 
we were able to make most use of available additional resources. In this case, we showed that both additional resources, the data for POS tagging as well as the named entity-annotated corpus, were beneficial for the translation quality. It is worth mentioning that this was achieved using corpora from a different domain, i.g. spoken TED talks versus written style. 
Furthermore, these corpora were significantly smaller than the available parallel data. Finally, the amount of parameter sharing defined by the architecture of the model has less influence on the final performance. Although, the best performance on both tasks was achieved with a model sharing only the encoder between the tasks.

In this work, the performance of machine translation task was improved by adopting multi-task training with other source language NLP tasks. In future work, we will also investigate methods to include target-language NLP tasks into the joint framework.

\section*{Acknowledgments}
The project leading to this application has received funding from the European Union's Horizon 2020 research and innovation programme under grant agreement n$^\circ$ 645452.
This work was supported by the Carl-Zeiss-Stiftung.

\bibliography{emnlp2017}
\bibliographystyle{emnlp_natbib}

\end{document}